\documentclass[letterpaper]{article} % DO NOT CHANGE THIS
\usepackage{aaai24}  % DO NOT CHANGE THIS
\usepackage{times}  % DO NOT CHANGE THIS
\usepackage{helvet}  % DO NOT CHANGE THIS
\usepackage{courier}  % DO NOT CHANGE THIS
\usepackage[hyphens]{url}  % DO NOT CHANGE THIS
\usepackage{natbib}  % DO NOT CHANGE THIS AND DO NOT ADD ANY OPTIONS TO IT
\usepackage{caption} % DO NOT CHANGE THIS AND DO NOT ADD ANY OPTIONS TO IT
\usepackage{xcolor}
\usepackage{pifont}
\usepackage{tikz}
\usepackage{fontawesome5}
\usepackage{framed}
\usepackage{times}  % DO NOT CHANGE THIS
\usepackage{helvet}  % DO NOT CHANGE THIS
\usepackage{courier}  % DO NOT CHANGE THIS
\usepackage[hyphens]{url}  % DO NOT CHANGE THIS
\usepackage{graphicx} % DO NOT CHANGE THIS
\usepackage{natbib}  % DO NOT CHANGE THIS AND DO NOT ADD ANY OPTIONS TO IT
\usepackage{caption} % DO NOT CHANGE THIS AND DO NOT ADD ANY OPTIONS TO IT
\usepackage{algorithm}
\usepackage{algorithmicx}
\usepackage{algpseudocode}
\usepackage{amsmath}     % for advanced math symbols
\usepackage{amsfonts}    % for math fonts
\usepackage{graphicx}    % for including graphics

\usepackage{newfloat}
\usepackage{listings}

\definecolor{mylightblue}{RGB}{100,200,255}
\definecolor{deepgreen}{RGB}{75,175,120}
\definecolor{deepred}{RGB}{255,50,100}
\definecolor{brownyellow}{RGB}{250,189,115}

\DeclareCaptionStyle{ruled}{labelfont=normalfont,labelsep=colon,strut=off} % DO NOT CHANGE THIS
\floatstyle{ruled}
\newfloat{listing}{tb}{lst}{}
\floatname{listing}{Listing}
\title{Boosting Logical Reasoning in Large Language Models through a New Framework: The Graph of Thought}
\author{
    Bin Lei\textsuperscript{\rm 1},
    Pei-Hung Lin\textsuperscript{\rm 2},
    Liao Chunhua\textsuperscript{\rm 2},
    Caiwen Ding\textsuperscript{\rm 1}
}
\affiliations{
    \textsuperscript{\rm 1}University of Connecticut\\
    \textsuperscript{\rm 2}Lawrence Livermore National Laboratory\\
    bin.lei@uconn.edu,
    lin32@llnl.gov,
    liao6@llnl.gov,
    caiwen.ding@uconn.edu
}

\usepackage{bibentry}
\begin{document}

\maketitle

\begin{abstract}
Recent advancements in large-scale models, such as GPT-4, have showcased remarkable capabilities in addressing standard queries. However, when facing complex problems that require multi-step logical reasoning, their accuracy dramatically decreases. Current research has explored the realm of \textit{prompting engineering} to bolster the inferential capacities of these models. Our paper unveils a pioneering prompting technique, dubbed \textit{Graph of Thoughts (GoT)}. Through testing on a trio of escalating challenges: the 24-point game, resolution of high-degree polynomial equations, and derivation of formulas for recursive sequences, our method outperformed GPT-4, achieving accuracy improvements of $89.7\%$, $86\%$, and $56\%$ for each respective task. Moreover, when juxtaposed with the state-of-the-art (SOTA) prompting method, \textit{Tree of Thought (ToT)}, our approach registered an average accuracy boost of $23\%$, $24\%$, and $15\%$.
\end{abstract}

\section{Introduction}
While large language models (LLMs) have revolutionized productivity with their ability to address a plethora of basic queries~\cite{dwivedi2023so}, rooted in their extensive knowledge, they still grapple with inherent limitations in cognitive \cite{chen2021evaluating} and reasoning skills~\cite{sap2022neural}. This deficit becomes evident in tasks demanding multi-step considerations \cite{creswell2022selection,paranjape2023art,nye2021work,kojima2022large,shridhar2023distilling}, even when employing cutting-edge models like GPT-4~\cite{openai2023gpt4}. 

\begin{figure}[h]  
\centering 
\includegraphics[width=0.47\textwidth]{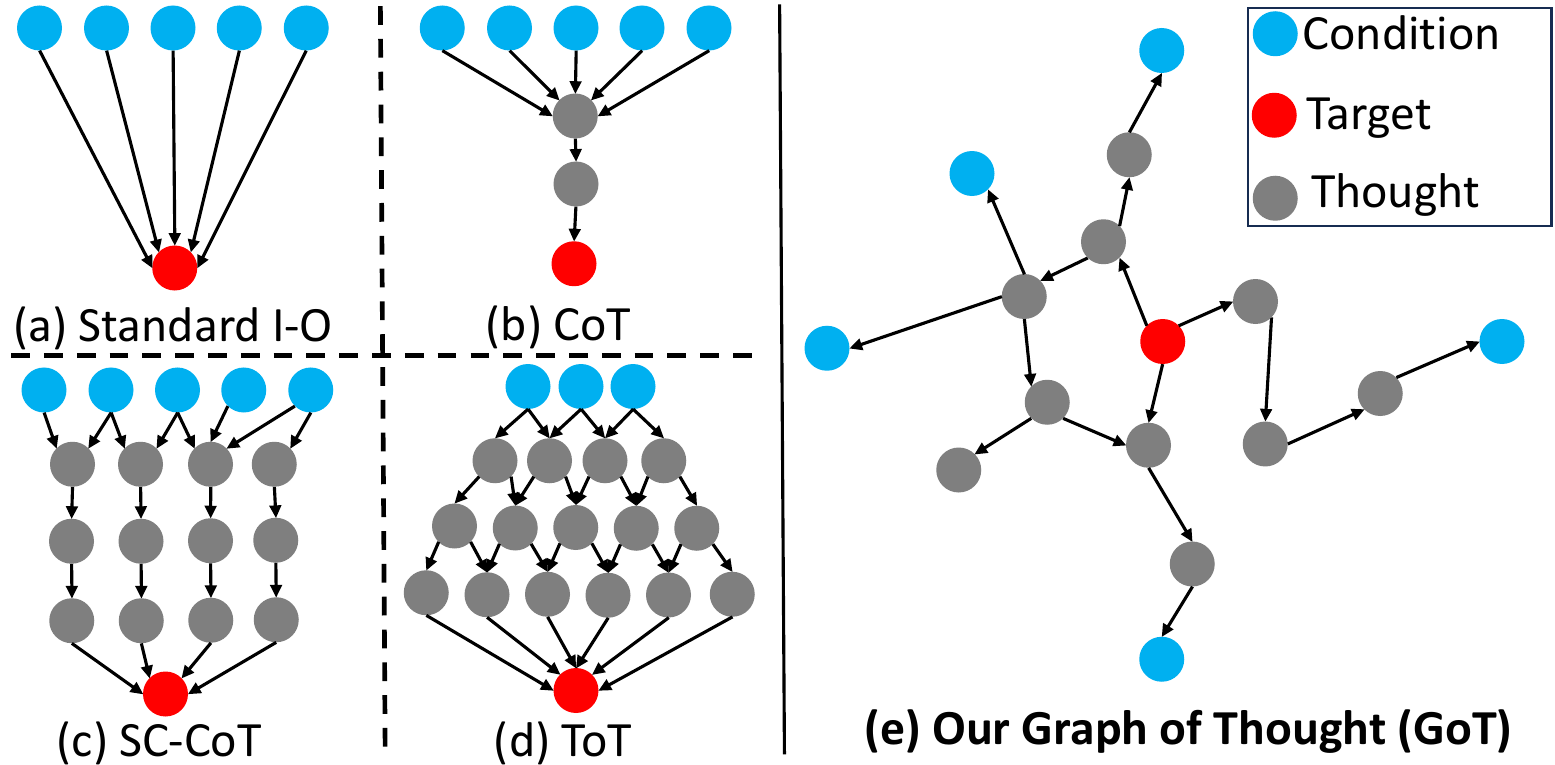} 
\caption{Comparison of various prompting approaches.}  
\label{fig:Introduction_figure}
\end{figure}

Many prior works have tried to address the weaker logical reasoning capabilities of LLMs using prompting engineering \cite{chang2023prompting,zhou2022large,strobelt2022interactive,jiang2022promptmaker,wu2022promptchainer} approaches, such as the \textit{Chain-of-Thought (CoT)}\cite{wei2022chain}, \textit{Self-Consistency of Chain-of-Thought SC-CoT} \cite{wang2022self}, and \textit{Tree-of-Thought (ToT)}~\cite{yao2023tree} methods. Among these, the  ToT method performs the best. It achieves an accuracy of $74\%$ in certain logical reasoning tasks like the 24-point game \cite{yao2023tree}, which is significantly higher than the default GPT-4's $7.3\%$ \cite{yao2023tree}. These four methods are
% respectively 
illustrated in (a) to (d) on the left side of Figure \ref{fig:Introduction_figure}.
% and will be further introduced in the following Related Work Section. 

Despite the success, these methods are still far away from practical usage. For instance, using the ToT method achieves a $74\%$ accuracy rate in the 24-point game, which still lags behind the human reasoning capabilities, with the human performance baseline in this game being approximately $98.5\%$ \cite{4nums2023}.
 \textit{Can we further enhance the reasoning capability of large models to achieve or even surpass human-level performance}? In light of this, this paper introduces a method named \textit{Graph of Thought (GoT)}.

Our approach is inspired by the emulation of human cognitive processes \cite{wang2010cognitive,kamijo2007interactive,estes2022handbook}. Let's consider one of the most renowned mathematical logic problem, the Goldbach's Conjecture \cite{wang2002goldbach,carbo2016study}, mathematicians do not attempt to enumerate all possible techniques and theorems. Instead, they reason backward from the conclusion \cite{carbo2016study,bamberg2003crystallographic,wang2022proof}. They identify promising avenues of research, and ascertain the essential foundational knowledge required to pursue a particular line of thought. Importantly, different lines of thought are not isolated; they are interconnected and collaboratively contribute towards forming the final solution \cite{granville2007refinements,oliveira2014empirical}. 

Consequently, diverging from the previously established works, this paper introduces a distinct problem-solving approach.
% Compared to the previous SOTA method (\textit{e.g., ToT}), our approach strengthens the connections between nodes and introduces a filtering mechanism, resulting in significant advancements in handling such complex tasks.
% Our methodology commences at the conclusion \textcolor{red}{(no idea what "commences at the conclusion" mean. Rewrite.)} and, through a graph-based technique, constructs various pathways culminating in that final outcome. Moreover, it strengthens the interlinkages between these routes \textcolor{red}{(No idea what this sentence is about. Delete or rewrite)}.
It mainly contributes in three aspects:

\begin{enumerate}
    \item Introduction of a novel \textbf{graph structure} to enhance the connections between different lines of thought (nodes).
    \item Implementation of a \textbf{checking mechanism} to ensure the accuracy of the connections between different lines of thought.
    \item Proposal of a new \textbf{graph updating} method, designed to facilitate rapid iteration for further reasoning.
\end{enumerate}

These points will be further detailed in the \textit{Graph of Thought} Section.

On the right side of Figure \ref{fig:Introduction_figure}, we present the
 sturcture of our Graph. It distinguishes itself from previous ToT or CoT in the following two aspects:

% \textcolor{red}{(contribution is too shallow. Need to bring more content from Section "GOT": Our design contains three key innovations)}
\begin{itemize}
    \item Our prompting approach initiates from the outcome rather than the conditions.
    \item Graph-based structure eliminate any inherent hierarchy among the intermediate nodes. This design allows for potential relationships between any pair of intermediate nodes.
\end{itemize}
We test our method on three logic reasoning tasks of increasing difficulty. The accuracy of our method surpass the current SOTA in all cases. In the 24-point game, its accuracy reaches $97\%$, which is $23$ percentage points higher than the current SOTA, approaching human reasoning levels ($98.5\%$ \cite{4nums2023}).

\section{Related Work}
\textit{I-O} Prompting:
The most prevalent prompting method is the Input-Output prompting. In this approach, one provides the conditions to the large model, which then produces answers following a token-level, left-to-right decision-making process \cite{zhang2022automatic}. This method is the default mode utilized by GPT-4.

\textit{CoT} Prompting \cite{wei2022chain}:
CoT prompting aims to guide the model in generating coherent text by establishing logical continuity. It is based on the assumption that by progressively expanding and supplementing chains of viewpoints and arguments, the model can generate more coherent and reasoning-based outputs. This method encourages the model to follow a sequential line of thought while answering questions, ensuring that the generated text is logically connected and coherent.

\textit{SC-CoT} Prompting \cite{wang2022self}:
SC-CoT prompting is an extension of the Chain-of-Thought method. It requires the model to maintain self-consistency and logical coherence while generating text. In other words, the generated content should be internally consistent and semantically connected. This consistency and logical flow can be ensured by leveraging large models to vote or score, followed by a selection process. By emphasizing the self-consistency of the model's output, this prompting method reduces logical errors and inconsistencies, thereby improving the quality of the generated text.

\textit{ToT} Prompting \cite{yao2023tree}:
ToT prompting employs structured prompts to assist the model in generating hierarchical and structured text. This method utilizes a tree-like structure to represent the relationships between different concepts, which serves as input prompts. The model can organize and reason about the text based on the structure of the tree, resulting in more accurate and structured answers. Tree of Thought also employs a voting or scoring mechanism to filter the generated results, thereby not only reducing errors but also decreasing computational costs. Tree-of-Thought prompting aims to provide richer semantic representation and improved logical organization, enhancing the model's reasoning capabilities and generation quality.

These are the four prompting methods, each offering guidance and constraints at different levels to facilitate language models in generating accurate, coherent, and reasoning-based text. 

\section{Graph of Thought}

Our design contains three key innovations:
\begin{enumerate}
\item \textbf{Graph Structure}: By constructing thought graphs, we model various aspects and concepts of the problem as nodes, while their relationships and connections are represented by edges. This graphical representation facilitates the LLM in capturing and comprehending complex logical relationships, thereby improving its reasoning capabilities and accuracy in answering questions.

\item \textbf{Inspection Mechanism}: We address the challenge of reducing errors in LLMs by implementing a rechecking process for potentially correct results. In this process, we provide a more robust estimation of result accuracy. By calculating the confidence or probability of different candidate answers, we can effectively assess and weigh their reliability. This allows us to select the final answer based on the probabilities associated with each candidate result. Our re-evaluation mechanism is executed using our Checker function, which offers greater precision compared to traditional scoring or voting systems. This enhanced accuracy arises primarily because: i) we filter for a single, optimal result rather than selecting several decent outcomes, and ii) our checker function comprises multiple linked inspectors, ensuring a more rigorous review.

\item \textbf{Convenient Graph Updating}: Throughout the graph traversal process, thoughts that pass the Checker function are continuously added to our condition sequence. For complex problems that may not be resolved in a single graph traversal, there's no need to keep track of all previously traversed paths. We can simply re-input our updated condition sequence. This significantly reduces potential redundant reasoning. A prime example of this can be seen in our experimental section's third task, with the graph update process illustrated in Figure \ref{fig:Numerical_problem}.
\end{enumerate}

Our first key point is detailed in the \textit{Graph Construction} Section. Here, we delve into the process of creating our graph, its distinct features, and how to interpret it, elucidating with a toy example. The second and third focal points are elaborated in the \textit{Graph Updating and Path Finding} Section. This section encompasses a comparative probabilistic analysis between our Checker function and traditional scoring mechanisms, as well as our graph updating algorithm.

\subsection{Graph Construction}\label{Graph_of_Thought}
We first need to establish a directed cognitive graph, which possesses the following characteristics:
\begin{itemize}
    \item The graph's creation originates from our final target.
    \item The graph contains some AND-Crossroad nodes. These AND-Crossroad nodes can only be returned if all paths from the intersection are unobstructed.
    \item The graph includes some Condition nodes, which can return to any node as long as there's a path between them.
    \item For the remaining nodes, as long as there is an unobstructed path from the node, it can be returned.
    \item Whether there's an obstacle on a path is determined by the Checker function.
    \item If a path can begin with certain condition nodes and return uninterrupted to the final target, then it is considered a valid path. This path can be used as the final output.
\end{itemize}

\begin{figure}[h]  
\centering 
\includegraphics[width=0.46\textwidth]{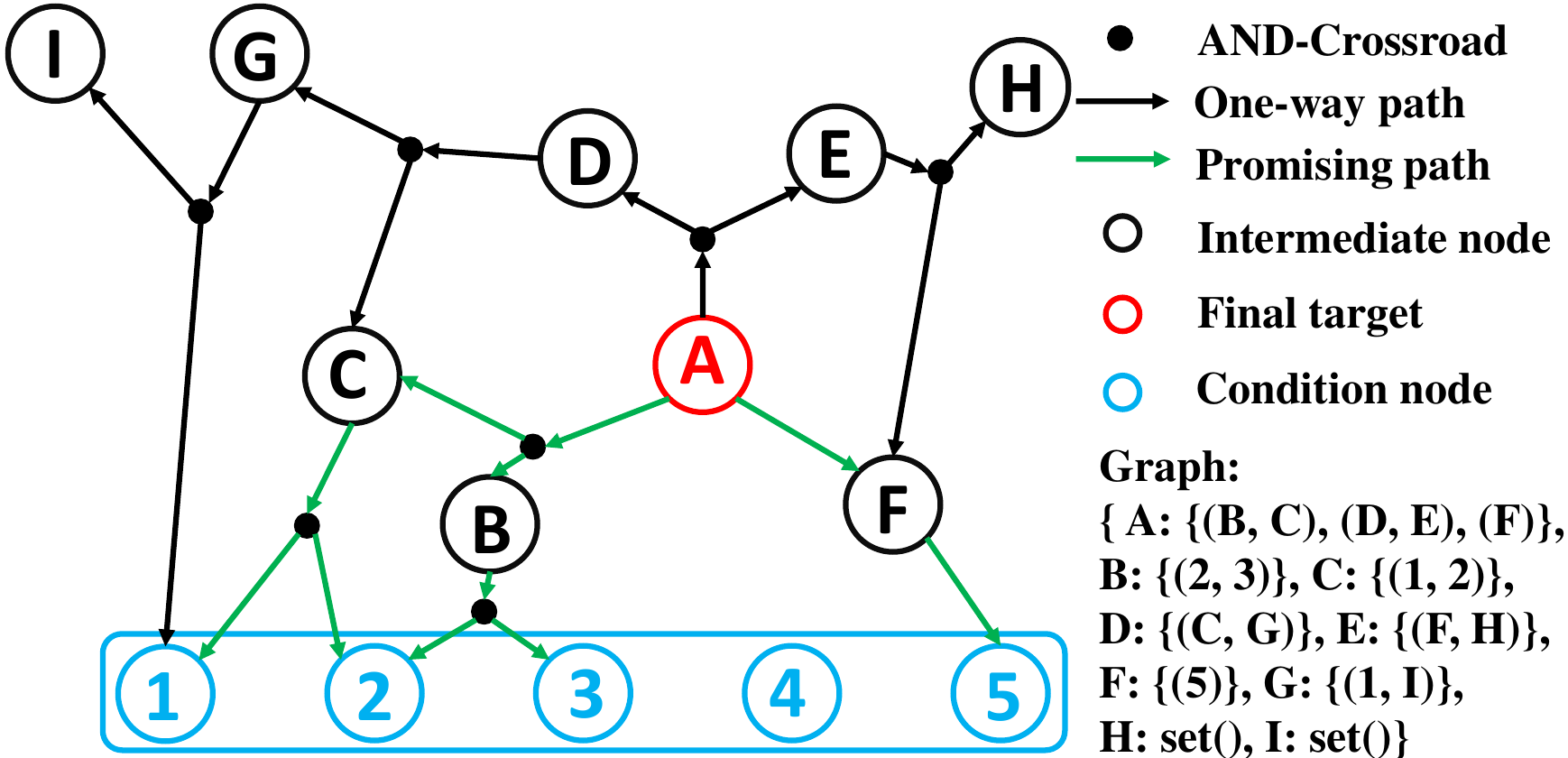} 
\caption{A toy example for the Graph of thought}  
\label{fig:Graph_construction}
\end{figure}

A toy example is illustrated in Figure~\ref{fig:Graph_construction}. In this instance, the graph can be represented as: 
\{ A: \{(B, C), (D, E), (F)\},
B: \{(2, 3)\},
C: \{(1, 2)\},
D: \{(C, G)\},
E: \{(F, H)\},
F: \{(5)\},
G: \{(1, I)\},
H: \{\},
I: \{\}\}.
Our ultimate destination is Node A. The creation of the graph begins here. Two promising paths, namely \([(1,2), (2,3), (C,B), (A)]\) and \([(5), (F), (A)]\), are highlighted in green. In this example, there are five condition nodes: Node 1, Node 2, Node 3, Node 4, and Node 5. These nodes can return to any AND-Crossroad or intermediate node without any prerequisite conditions. Node 5 leads to Node F, hence Node F is considered as a returnable node. Both Node 1 and Node 2 are returnable nodes, therefore, they can pass through the AND-crossroad and return to Node C. Similarly, Nodes B and A can be reached.

The construction of the graph is according to the Algorithm \ref{alg:create_graph}.
Our algorithm adopts a depth-first traversal approach to recursively create the mental graph. In this process, we call the LLM twice. The first call to LLM is to find paths based on the known node information, and the second call is to find new nodes that can reach the current node. We can manually set the number of searches to control the size of the graph, or continue searching until all new nodes are included in the Condition nodes or no new nodes can be found.

\begin{algorithm}[h]
\caption{Create the Graph}
\label{alg:create_graph}
\textbf{INPUT}: Conditions $Cs$, Nodes $Ns$, Graph $G$\\
\textbf{OUTPUT}: Updated Graph $G$
\begin{algorithmic}[1]
\Function{Create\_Graph}{$Cs, Ns, G$}
    \For{$N$ \textbf{in} $Ns$}
        \If{$N$ \textbf{in} $Cs$}
            \State \textbf{continue} 
        \EndIf
        \State $N \rightarrow$ LLM $\rightarrow paths$  \Comment{Create paths by LLM}
        \State $G[T] \gets \{\}$ \Comment{Create new node}
        \For{$path$ \textbf{in} $paths$}
            \State $path \rightarrow$ LLM $\rightarrow ns$ \Comment{$ns$ : new nodes}
            \State $G[T]$.add(\textbf{tuple}($ns$)) \Comment{update the Graph}
            \State \Call{Create\_Graph}{$Cs$, $ns$, $G$}
        \EndFor
    \EndFor
\EndFunction
\end{algorithmic}
\end{algorithm}

\subsection{Graph Updating and Path Finding}
In our Graph, conditions play a crucial role in achieving the final result. The so-called graph updating is essentially a process of continually adding intermediate nodes to the sequence of condition nodes. Updating the graph serves two main purposes:

\begin{enumerate}
\item \textbf{Facilitating Graph Reuse:} Suppose we can not find the correct path when first creating the graph and conducting the path finding operation. When recreating the graph and performing the path finding operation based on this, we only need to take the condition nodes updated based on this graph as the new condition nodes. Without the need to record all the paths in this graph, we can effectively avoid recreating the graph.

\item \textbf{Efficiency in Path Finding:} In the process of path finding, there is no need to recursively look at all paths. We can search directly from the condition nodes, thereby improving efficiency.
\end{enumerate}

Our graph updating algorithm is shown in the Algorithm \ref{alg:Update_Graph}. We traverse all nodes in the graph, performing depth-first search with each node as the root node. If a certain node is identified as a returnable node, we add this node to the original list of condition nodes. In our thought graph, each conditional node will eventually be used to calculate the final result. Therefore, we need to be extremely careful when adding intermediate nodes to the list of conditional nodes. To maximize accuracy, we have implemented a multiple verification mechanism. To deem an intermediate node as returnable, it must not only have a path connected to condition nodes, but every such path must also be unobstructed. Only paths that successfully pass through our Checker function can be considered unobstructed. 

The Checker function is also shown in the Algorithm \ref{alg:Update_Graph}. We use the information of this path and its two end nodes as inputs for the Checker function. Using LLM, we repeatedly judge whether we can correctly obtain the child node through this path. If each LLM determination confirms that the path is indeed feasible, the checker returns True. Otherwise, it returns False.

Compared to previous scoring mechanisms, our approach leans more towards inspection rather than selection. This practice can significantly reduce the chance of misjudgment, as analyzed below.

For the original scoring function $F_{scoring}$:
\begin{align*}
F_{\text{scoring}}(P_{\text{LLM}}, C)(s) &= 1[s_{max}~\geq~s~\geq~s_{t}] 
\end{align*}
Where $\text{score}~s~\sim~P_{\text{LLM}}(s|C)$, $P_{LLM}$ is the probability distribution of the LLM, $C$ represents the current set of conditions, $s_{max}$ is the maximum score, $s_{t}$ is the selection threshold score.

The probability of selecting a certain path as a promising path in the scoring mechanism is:  
\vspace{-0.2cm}
\begin{align*}
P_{F_{scoring} = 1}  &= \sum_{s=s_{t}}^{s_{max}} P_{LLM}(s|C) 
\end{align*}
For our inspection function $F_{inspection}$:
\begin{align*}
F_{\text{inspection}}(P_{LLM}, C)(s) &= 1[r = True]
\end{align*}
Where $\text{result}~r~\sim~P_{checker}(r|C)$, and each check is independent and identically distributed. 

The probability of selecting a certain path as a correct path in our inspections mechanism is:
\begin{align*}
P_{F_{inspection} = 1}  &= P_{checker}(True|C) = (P_{LLM}(s_{max}|C))^n
\end{align*}
Where $n$ is the number of inspectors. 

The comparison between the two selection methods is as follows:
\vspace{-0.3cm}
\begin{align*}
(P_{LLM}(s_{max}|C))^n < P_{LLM}(s_{max}|C) < \sum_{s=s_{t}}^{s_{max}} P_{LLM}(s|C)
\end{align*}
% \vspace{-0.3cm}
Evidently, the selection criteria of our inspection method are more stringent. This is one of the key reasons contributing to the accuracy improvement. The specific comparison of accuracy is demonstrated in our \textit{Experiment} Section.

% \textcolor{red}{therefore better than tree}

Path finding represents the final step in our algorithm, the primary objective of which is to utilize the conditions to compute our ultimate results. After the graph has been updated, promising intermediate nodes have already been incorporated into the conditions list, thus simplifying the path finding process. This operation can be accomplished with a single loop, which we will not belabor here.

Notably, due to our stringent selection criteria, there may be instances when we need to rebuild the graph to locate suitable paths. In these scenarios, by utilizing the updated conditions list, we can retain information from all previous graphs. This approach allows us to avoid duplication of paths without maintaining the prior graph information. A good example is the third task in our \textit{Experiment} Section as shown in Figure \ref{fig:Numerical_problem}.

\begin{algorithm}[h]
\caption{Update the Graph}
\label{alg:Update_Graph}
\textbf{INPUT}: Graph $G$, Conditions $Cs$, distance $D$, number of inspectors $n$\\
\textbf{OUTPUT}: Updated Graph $G$
\begin{algorithmic}[1]
\Function{Update\_Graph}{$G, Cs, D$}
    \If{$D == 0$}
    \State \textbf{return} $G$
    \EndIf
    \State $new\_Cs \gets list()$ \Comment{Initialize new conditions}
    \For{$node \in G$}
        \For{$path \in G[node]$}
             \If{\textbf{all} $items$ \textbf{in} $path$ \textbf{are in} $Cs$ \textbf{AND} \Call{Checker}{$node,path,n$}}
            \State $new\_Cs.\text{append}(node)$
            \State \textbf{break} 
            \EndIf
        \EndFor
    \EndFor
    \State $Cs \gets Cs \cup new\_Cs$ \Comment{Update the condition set}
    \State $G \gets G \setminus new\_Cs$ \Comment{Update the graph }
    \State \Call{Update\_Graph}{$G, Cs, D-1$}
\EndFunction
\Statex
\Function{Checker}{$node,path,n$} \Comment{Multiple verification}
    \For{$i \in \text{range}(1, n+1)$}
        \State $result(True/False) \gets LLM(node,path)$ 
        \If{$\neg result$}
        \State \textbf{return False}
        \EndIf
    \EndFor
    \State \textbf{return} True
\EndFunction
\end{algorithmic}
\end{algorithm}

\section{Experiment} \label{sec:experiment}
We conduct three distinct experimental sets to validate the effectiveness of our methodology: the 24-point game, the resolution of high-order equations, and the computation of arithmetic sequence formulas. The complexity of these tasks escalates in the given order. For each experiment, we begin by illustrating how the LLM leverages our method to generate results, using a specific example for clarity. In detailing this procedure, key outcomes linked to pivotal intermediate nodes crafted by the LLM were accentuated using \fbox{~~~~}. All experiments were executed using the Chat Completion mode of GPT-4 between July 1st and July 31st, 2023, with a set sampling temperature of 
$0.7$.

\subsection{24-Point Game}
The rule of this game is to calculate 24 using four given numbers and the operations of addition, subtraction, multiplication, and division. Each number must be used once and only once. Our dataset is from \cite{4nums2023}

Here is an example problem: 
\begin{framed}
Use 6, 10, 12, 13 and basic arithmetic operations (+ - * /) to obtain 24. Each number must be used once and only once.
\end{framed}

\begin{figure}[h]  
\centering 
% \begin{tikzpicture}
%     % Black ellipse
%     \draw [black] (0,0) ellipse [x radius=1cm, y radius=0.5cm];
    
%     % Red ellipse
%     \draw [red] (3,0) ellipse [x radius=1cm, y radius=0.5cm];
    
%     % Blue ellipse
%     \draw [blue] (6,0) ellipse [x radius=1cm, y radius=0.5cm];
% \end{tikzpicture}
\includegraphics[width=0.46\textwidth]{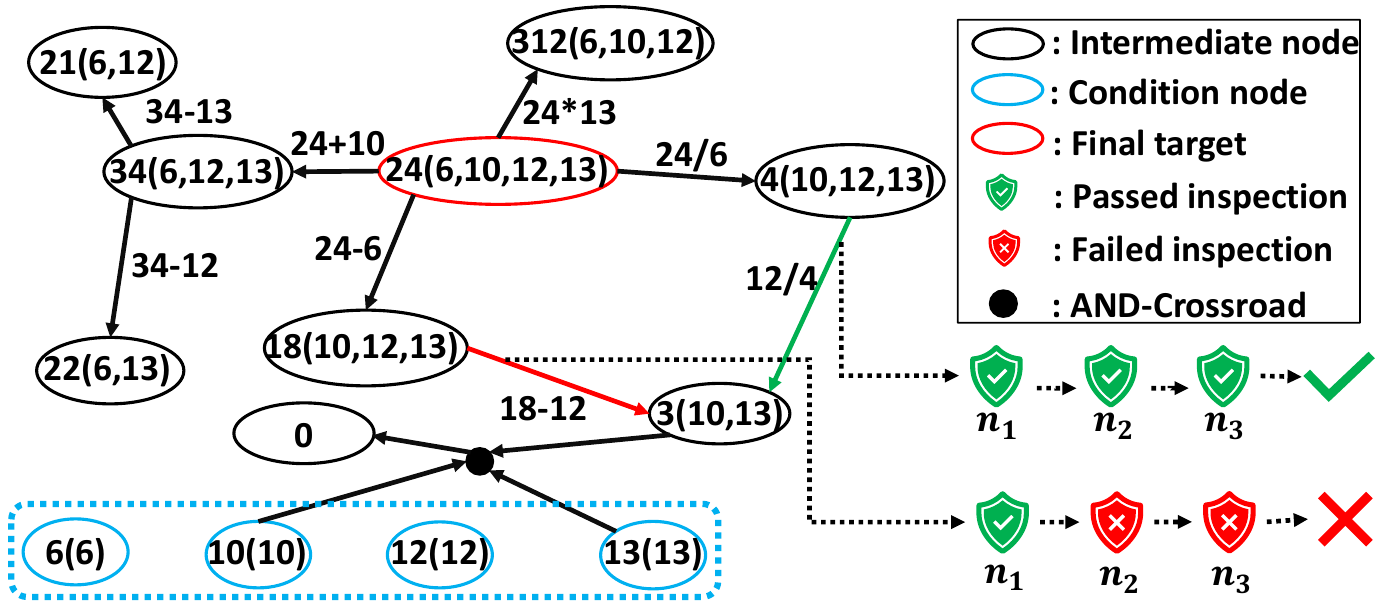} 
\caption{An example of GoT in 24-points game. $n_1$,$n_2$,$n_3$: Inspectors $1$,$2$,$3$.} 
\label{fig:24_game}
\end{figure}

Therefore, using such problems to test multi-step logical reasoning capabilities is a good choice. Figure \ref{fig:24_game} illustrates the process of solving this problem using our GoT method.

The number outside the bracket in the circle represents the current value, and the number inside the bracket represents the numbers that have not been used yet. The equations beside the paths indicate the specific computational steps of that path.  

Every path needs to be checked by the Checker function.
The inspection process for two paths is shown in the figure. In each Checker function, several inspectors are connected in series. The final Checker function only returns True if all inspectors return True. If the final result returns False, then the path cannot be traversed. As shown in the figure: The operation \fbox{$12/4=3$} generated by LLM passes the Checker function and is traversable, while \fbox{$18-12=3$} fails and is not traversable. 
Calculate starting from the final result, passing through traversable paths, aided by condition nodes, and reaching the endpoint via AND-Crossroad. Therefore, the correct answer to this problem is:~\fbox{$13-10=3$; $12/3=4$; $4*6=24$}.

The same 24-points game is also conducted in the ToT study. We compare our results with theirs and presented the outcomes in Table~\ref{tab:24 game}. In the CoT-SC experiment, $k=100$ denotes the success rate calculated using the best of $k$ samples. In the ToT experiment, the authors conduct a breadth-first search, where $b$ in the table stands for 'breadth'. In the GoT experiment, $n$ denotes the number of inspectors in the Checker function.
\begin{table}[htbp]
  \centering
  \caption{GoT vs. Other Methods in 24-Point Game}
    \begin{tabular}{cc}
    \hline
    \textbf{Method} & \textbf{Accuracy} \\
    \hline
    IO~\cite{yao2023tree}    & 7.3\% \\
    CoT~\cite{yao2023tree}   & 4.0\% \\
    CoT-SC (k = 100)~\cite{yao2023tree} & 9.0\% \\
    ToT (b = 1)~\cite{yao2023tree} & 45\% \\
    ToT (b = 5)~\cite{yao2023tree} & 74\% \\
    \hline
    \textbf{GoT (n = 0)} & \textbf{77\%} \\
    \textbf{GoT (n = 1)} & \textbf{85\%} \\
    \textbf{GoT (n = 3)} & \textbf{93\%} \\
    \textbf{GoT (n = 5)} & \textbf{97\%} \\
    \hline
    \end{tabular}%
  \label{tab:24 game}%
\end{table}%

The results show that even when the number of inspectors $n = 0$, our accuracy in this game surpasses that of the Tree-of-Thought prompting method when $b = 5$. As the number of inspectors increases, more errors in the intermediate computation process are corrected, gradually improving the accuracy. When $n = 3$, our accuracy rate reaches $93\%$, which is a $\mathbf{\times 12.73}$ increase compared to the initial standard I-O prompting. When $n = 5$, our accuracy rate reaches $97\%$, which is a $\mathbf{23\%}$ higher compared to the ToT when $b = 5$.

\subsection{Solving High-Degree Polynomial Equations}
Next, we increase the difficulty of the task by deploying GoT to tackle higher-degree equations. While standard formulas exist for the roots of low-degree equations, their counterparts for higher degrees demand more intricate solutions, such as Newton's method or the Durand-Kerner method. Consider equations like $x^6 + 3x^4 - 2018x^3 + 3x^2 + 1 = 0$ or $x^4 - 3x^3 + 3x + 1 = 0$; they don't lend themselves to straightforward solutions. We apply our GoT approach to this problem and compared its performance with other methods.

Our dataset is derived from the Mathematics Dataset \cite{saxton2019analysing}. 
Here is an example problem:
\begin{framed}
Solve the equation: $3x^4 -69x^3 + 1284x^2 - 4212x -3888 = 0$
\end{framed}

\begin{figure}[h]  
\centering 
\includegraphics[width=0.46\textwidth]{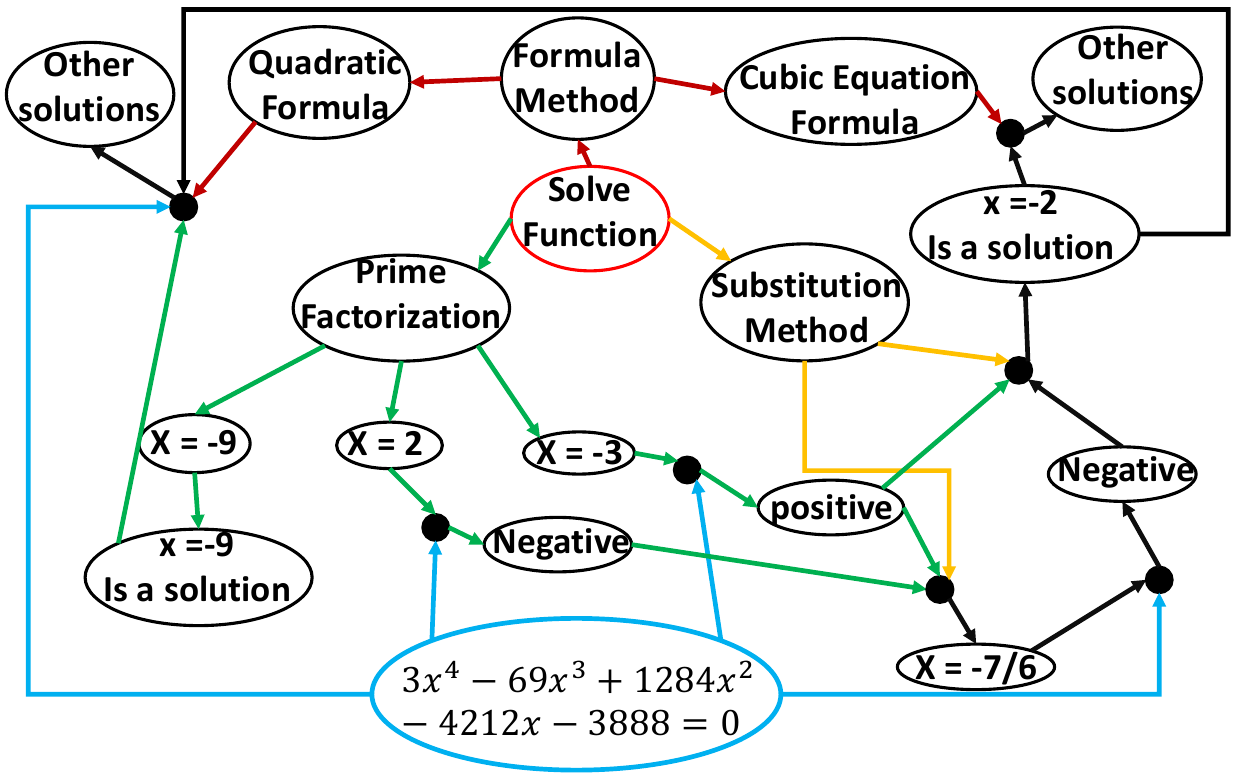} 
\caption{An example of GoT in Solving Polynomial Functions.
Black oval: Intermediate node; 
Red oval: Final target; 
Blue oval: Condition nodes;
Black dot: AND-Crossroad;
Green arrow: Prime Factorization related path;
Blue arrow: Condition related path;
Red arrow: Formula Method related path;
Yellow arrow: Substitution Method related path.}
\label{fig:polynomial}
\end{figure}

This problem is more complex than the previous 24-point problem. Figure \ref{fig:polynomial} illustrates the process of our GoT method in addressing this kind of problem.

We start with the goal of solving the equation, and in this example, the LLM provides us with three possible approaches: the formula method, Prime factoring, and numerical substitution. In the figure, we mark them with red arrows, green arrows, and yellow arrows respectively. We list some examples of the LLM's attempts, such as for the factoring method, where the LLM tried \fbox{-9, 2, and -3}. Among these, \fbox{-9 is a solution} to the equation, and when substituting \fbox{\(x = 2\) and \(x = -3\)} into the left side of the original equation, the values of the equation are negative and integer, respectively. Therefore, the LLM suggests using the numerical method to try some potential solutions between these two numbers. 
First, it tries \fbox{\(x = -\frac{7}{6}\)}, but when this value of \(x\) is substituted into the left side of the original equation, the value is still negative. Afterward, the LLM suggests continuing to try possible factors between \fbox{\(-\frac{7}{6}\) and -3}, ultimately finding that \fbox{\(x = -2\) is a solution} to the original equation.
At this point, with the help of the large model, we have found two solutions to the equation, \fbox{\(x = -9\) and \(x = -2\)}. The large model then suggests that we substitute these solutions back into the original equation, and then use the formulas for solving \fbox{cubic or quadratic equations} to find the remaining solutions.

In this experiment, we test the dataset using IO, CoT, and ToT methods respectively and compared them with our GoT method. 
\begin{itemize}
    \item For the \textbf{IO} method, the prompt reads: \texttt{"Please help me to solve the following equation: "} + \textit{problem}.
    
    \item In the \textbf{CoT} method, the prompt structure is: \texttt{"I want to solve "} + \textit{problem} + \texttt{"We have already tried "} + \textit{already tried steps} + \texttt{"What is the best next step?"} After the model suggests a subsequent step, its response is incorporated into the \textit{already tried steps}. The model is then re-queried until it either provides a perceived correct answer or indicates uncertainty. This prompt employs a 3-shots approach.
    
    \item For the \textbf{ToT} method, the prompt formulation is: \texttt{"I want to solve "} + \textit{problem} + \texttt{"We have previously attempted "} + \textit{already tried steps} + \texttt{"What should I do next?"}. Once the model delineates the forthcoming steps, each step is considered a node. We then systematically integrate the prompts from each node into the \textit{already tried steps}, limiting the traversal depth to 5. This prompt is designed with a 5-shots strategy.
    
\end{itemize}

In this experiment, we find that the accuracy of the large model is too low in the step of substituting values into the original equation to verify. Therefore, we not only tries to use our Checker function but also attempt to directly open calculator functionality to the large model. The comparison of various results is shown in the Table \ref{tab:polynomial}.

\begin{table}[htbp]
  \centering
  \caption{GoT vs. Other Methods in Solving Polynomial Equations}
    \begin{tabular}{cc}
    \hline
    \textbf{Method} & \textbf{Accuracy} \\
    \hline
    IO    & 3.0\% \\
    CoT   & 21\% \\
    ToT (b = 5) & 25\% \\
    ToT (with Calculator) & 65\% \\
    \hline
    \textbf{GoT (n = 0)} & \textbf{31\%} \\
    \textbf{GoT (n = 1)} & \textbf{45\%} \\
    \textbf{GoT (n = 5)} & \textbf{73\%} \\
    \textbf{GoT (with Calculator)} & \textbf{89\%} \\
    \hline
    \end{tabular}%
  \label{tab:polynomial}%
\end{table}%

From the table, we can see that although the large model's algebraic calculation ability is quite worrying (the accuracy rate of the ToT method surprisingly increased by 40\% after providing a calculator), our Checker function plays a very good error correction role. When the number of inspectors $n$ is set to $5$, the accuracy of our GoT method has surpassed that of ToT with a calculator. Moreover, if a calculator is available to the GoT method, its accuracy can reach $89\%$.

\subsection{Deriving Formulas for Recursive Sequences}
The final set of experiments aimed to evaluate the LLM's reasoning abilities in a more challenging scenario: deriving recurrence relations for sequences. We have collected sequence-related problems from mathematics competitions spanning nearly 20 years, and our dataset is provided in the supplementary material.

Here is an example problem: 
\begin{framed}
In the sequence \(a_n\), \(a_1 = 1\), and for \(n \geq 1\), \(a_{n + 1} = (1 + \frac{1}{n}) \cdot a_n + \frac{n + 1}{2^n}\). Find the general formula for the sequence \(\{a(n)\}\).
\end{framed}

\begin{figure*}[h]  
\centering 
\includegraphics[width=1\textwidth]{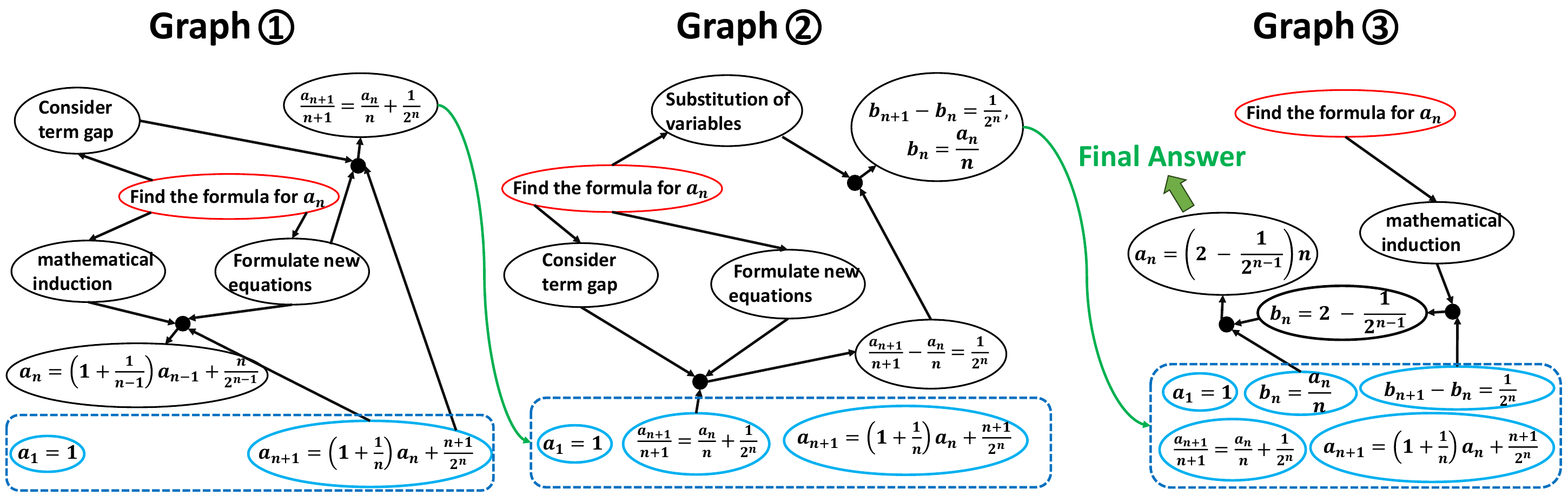} 
\caption{An example of GoT in Deriving the recurrence formula. Graph \ding{192}: The first graph traversal; Graph \ding{193}: The second graph traversal; Graph \ding{194}: The third graph traversal; Black oval: Intermediate node; 
Red oval: Final target; 
Blue oval: Condition nodes;
Black dot: AND-Crossroad;
Green arrow: The conversions from intermediate node to condition node.}
\label{fig:Numerical_problem}
\end{figure*}

 For complex mathematical problems like this, a single round of simple graph traversal search is often insufficient to provide solutions. Multiple rounds of graph updates are needed, continually supplementing known conditions, in an attempt to obtain the eventual correct outcome. The Figure \ref{fig:Numerical_problem} below demonstrates an example of our approach to solving such problems.

In this example, we go through three rounds of graph traversal. After each traversal, new information is added to the condition sequence. In Graph \ding{192}, the large model provides three possible directions:
\begin{enumerate}
    \item Consider using mathematical induction.
    \item Computing the difference between adjacent terms.
    \item Try to construct a new equation using existing conditions.
\end{enumerate}
Combining 2 and 3, along with the original equation in the condition sequence 
\fbox{$a_{n + 1} = \left(1 + \frac{1}{n}\right) \cdot a_n + \frac{n + 1}{2^n}$}
the model derives a new equation: 
\fbox{$\frac{a_{n+1}}{{n+1}} = \frac{a_n}{n} + \frac{1}{2^n}$}
This equation successfully pass our verification mechanism and is added to the condition sequence after the first round of graph traversal.

In Graph \ding{193}, due to changes in the condition sequence, the output of the large model also differs from the first round of graph traversal. This time, it still provides three possible solutions: 
\begin{enumerate}
    \item Try to replace the existing variables.
    \item Calculating the difference between adjacent terms.
    \item Try to construct a new equation using existing conditions.
\end{enumerate}

Next, the large model, first by combining points 2 and 3, changes the form of the equation in the condition sequence from \fbox{$\frac{a_{n+1}}{n+1} = \frac{a_n}{n} + \frac{1}{2^n}$} to \fbox{$\frac{a_{n+1}}{n+1} - \frac{a_n}{n} = \frac{1}{2^n}$}. Subsequently, the large model combines this equation with point 1 and replaces $\frac{a_n}{n}$ with $b_n$, eventually obtaining \fbox{$b_{n+1} - b_n = \frac{1}{2^n}$}.

In Graph \ding{194}, Due to the sufficiency of the conditions, the large model directly suggests adopting the mathematical induction method to solve the problem. Through condition \fbox{$b_{n+1} - b_n = \frac{1}{2^n}$} and the mathematical induction method, the large model derives the result \fbox{$b_n = 2 - \frac{1}{2^{n-1}}$}. Subsequently, utilizing condition \fbox{$b_n = \frac{a_n}{n}$}, the model arrives at the result \fbox{$a_n = (2 - \frac{1}{2^{n-1}})\times n$}.

In this experiment, we test the dataset using IO, CoT, and ToT methods respectively and compare with GoT. 
\begin{itemize}
    \item In the \textbf{IO} method, the prompt formulation is: \texttt{"Please help me solve the following problem: "} + \textit{problem}.
    
    \item Using the \textbf{CoT} method, the prompt structure is: \texttt{"Given that"} + \textit{the condition part of the problem} + \texttt{"we want to determine "} + \textit{the question part of the problem} + \texttt{"What might be the next step?"} Once the model suggests a subsequent step, this response is appended to the condition segment of the problem. We then consult the model repeatedly until it offers a perceived correct solution or indicates uncertainty.
    
    \item For the \textbf{ToT} method, the prompt reads: \texttt{"Considering "} + \textit{the condition part of the problem} + \texttt{"we seek to find out"} + \textit{the question part of the problem} + \texttt{"What could be the potential next steps?"} Upon the model's recommendation of subsequent steps, its response is integrated into the condition segment of the problem. We then treate each recommended step as a distinct node, with a capped traversal depth of 5. We utilize a 5-shots prompting approach.
    
\end{itemize}

Subsequently, we provide the large model with two mathematical tools 
\begin{enumerate}
    \item Variable Transformations: \text{e.g.}, Transform $a_{n + 1} = \left(1 + \frac{1}{n}\right) \cdot a_n + \frac{n + 1}{2^n}$ to $a_{n} = \left(1 + \frac{1}{n-1}\right) \cdot a_{n-1} + \frac{n}{2^{n-1}}$.
    
    \item Formula Simplification: \text{e.g.}, Simplify $a_{n + 1} = \left(1 + \frac{1}{n}\right) \cdot a_n + \frac{n + 1}{2^n}$ to $\frac{a_{n+1}}{n+1} - \frac{a_n}{n} = \frac{1}{2^n}$.
\end{enumerate}

and conducted the tests again. The experimental results are shown in the Table \ref{tab:Recursive_sequence}. 

\begin{table}[htbp]
  \centering
  \caption{GoT vs. Other Methods in Solving Recursive Sequences}
    \begin{tabular}{cc}
    \hline
    \textbf{Method} & \textbf{Accuracy} \\
    \hline
    IO    & 1.0\% \\
    CoT   & 3.0\% \\
    ToT (b = 5) & 17\% \\
    ToT (with Mathematical
tools) & 42\% \\
    \hline
    \textbf{GoT (n = 0)} & \textbf{20\%} \\
    \textbf{GoT (n = 1)} & \textbf{31\%} \\
    \textbf{GoT (n = 5)} & \textbf{55\%} \\
    \textbf{GoT (with Mathematical tools)} & \textbf{57\%} \\
    \hline
    \end{tabular}%
  \label{tab:Recursive_sequence}%
\end{table}%

Using direct IO input, the accuracy is only 1\%. Testing with the CoT method yield a comparable accuracy of about 3\%. The methods ToT and GoT see improved accuracy rates, rising to 17\% and 20\% respectively. Before granting the large model access to mathematical tools, the best-performing model is the GoT model with number of inspections $n$ set to 5, achieving an accuracy of 55\%. After enabling the mathematical tools, the accuracy for the GoT model increases to 57\%, while the ToT model rise to 42\%.

\section{Conclusion}
In this study, we introduce \textit{Graph of Thoughts}, a novel prompting technique that significantly enhances the inferential capabilities of large language models. The experimental results on three tasks of increasing difficulty— the 24-point game, solving higher-degree equations, and deriving recursive formulas for sequences— demonstrate its superiority. When compare to the SOTA Language Model, GPT-4, our method boost accuracy by $89.7\%$, $86\%$, and $56\%$, respectively. Against the current best prompting strategy, our approach improve accuracy by $23\%$, $24\%$, and $15\%$ for the respective tasks. These findings underscore the significant advantage of our technique in assisting large models to accomplish complex multi-step logical reasoning tasks.

\bibliography{aaai24}

\begin{thebibliography}{29}
\providecommand{\natexlab}[1]{#1}

\bibitem[{4nums.com(2023)}]{4nums2023}
4nums.com. 2023.
\newblock {4nums.com - Best Math Game Online}.
\newblock Accessed: August 13, 2023.

\bibitem[{Bamberg, Cairns, and Kilminster(2003)}]{bamberg2003crystallographic}
Bamberg, J.; Cairns, G.; and Kilminster, D. 2003.
\newblock The crystallographic restriction, permutations, and Goldbach's
  conjecture.
\newblock \emph{The American mathematical monthly}, 110(3): 202--209.

\bibitem[{Carb{\'o}-Dorca(2016)}]{carbo2016study}
Carb{\'o}-Dorca, R. 2016.
\newblock A study on Goldbach conjecture.
\newblock \emph{Journal of Mathematical Chemistry}, 54: 1798--1809.

\bibitem[{Chang(2023)}]{chang2023prompting}
Chang, E.~Y. 2023.
\newblock Prompting large language models with the socratic method.
\newblock In \emph{2023 IEEE 13th Annual Computing and Communication Workshop
  and Conference (CCWC)}, 0351--0360. IEEE.

\bibitem[{Chen et~al.(2021)Chen, Tworek, Jun, Yuan, Pinto, Kaplan, Edwards,
  Burda, Joseph, Brockman et~al.}]{chen2021evaluating}
Chen, M.; Tworek, J.; Jun, H.; Yuan, Q.; Pinto, H. P. d.~O.; Kaplan, J.;
  Edwards, H.; Burda, Y.; Joseph, N.; Brockman, G.; et~al. 2021.
\newblock Evaluating large language models trained on code.
\newblock \emph{arXiv preprint arXiv:2107.03374}.

\bibitem[{Creswell, Shanahan, and Higgins(2022)}]{creswell2022selection}
Creswell, A.; Shanahan, M.; and Higgins, I. 2022.
\newblock Selection-inference: Exploiting large language models for
  interpretable logical reasoning.
\newblock \emph{arXiv preprint arXiv:2205.09712}.

\bibitem[{Dwivedi et~al.(2023)Dwivedi, Kshetri, Hughes, Slade, Jeyaraj, Kar,
  Baabdullah, Koohang, Raghavan, Ahuja et~al.}]{dwivedi2023so}
Dwivedi, Y.~K.; Kshetri, N.; Hughes, L.; Slade, E.~L.; Jeyaraj, A.; Kar, A.~K.;
  Baabdullah, A.~M.; Koohang, A.; Raghavan, V.; Ahuja, M.; et~al. 2023.
\newblock “So what if ChatGPT wrote it?” Multidisciplinary perspectives on
  opportunities, challenges and implications of generative conversational AI
  for research, practice and policy.
\newblock \emph{International Journal of Information Management}, 71: 102642.

\bibitem[{Estes(2022)}]{estes2022handbook}
Estes, W. 2022.
\newblock \emph{Handbook of learning and cognitive processes}.
\newblock Psychology Press.

\bibitem[{Granville(2007)}]{granville2007refinements}
Granville, A. 2007.
\newblock Refinements of Goldbach's conjecture, and the generalized Riemann
  hypothesis.
\newblock \emph{Functiones et Approximatio Commentarii Mathematici}, 37(1):
  159--173.

\bibitem[{Jiang et~al.(2022)Jiang, Olson, Toh, Molina, Donsbach, Terry, and
  Cai}]{jiang2022promptmaker}
Jiang, E.; Olson, K.; Toh, E.; Molina, A.; Donsbach, A.; Terry, M.; and Cai,
  C.~J. 2022.
\newblock Promptmaker: Prompt-based prototyping with large language models.
\newblock In \emph{CHI Conference on Human Factors in Computing Systems
  Extended Abstracts}, 1--8.

\bibitem[{Kamijo et~al.(2007)Kamijo, Nishihira, Higashiura, and
  Kuroiwa}]{kamijo2007interactive}
Kamijo, K.; Nishihira, Y.; Higashiura, T.; and Kuroiwa, K. 2007.
\newblock The interactive effect of exercise intensity and task difficulty on
  human cognitive processing.
\newblock \emph{International journal of psychophysiology}, 65(2): 114--121.

\bibitem[{Kojima et~al.(2022)Kojima, Gu, Reid, Matsuo, and
  Iwasawa}]{kojima2022large}
Kojima, T.; Gu, S.~S.; Reid, M.; Matsuo, Y.; and Iwasawa, Y. 2022.
\newblock Large language models are zero-shot reasoners.
\newblock \emph{Advances in neural information processing systems}, 35:
  22199--22213.

\bibitem[{Nye et~al.(2021)Nye, Andreassen, Gur-Ari, Michalewski, Austin,
  Bieber, Dohan, Lewkowycz, Bosma, Luan, Sutton, and Odena}]{nye2021work}
Nye, M.; Andreassen, A.~J.; Gur-Ari, G.; Michalewski, H.; Austin, J.; Bieber,
  D.; Dohan, D.; Lewkowycz, A.; Bosma, M.; Luan, D.; Sutton, C.; and Odena, A.
  2021.
\newblock Show Your Work: Scratchpads for Intermediate Computation with
  Language Models.
\newblock arXiv:2112.00114.

\bibitem[{Oliveira~e Silva, Herzog, and Pardi(2014)}]{oliveira2014empirical}
Oliveira~e Silva, T.; Herzog, S.; and Pardi, S. 2014.
\newblock Empirical verification of the even Goldbach conjecture and
  computation of prime gaps up.
\newblock \emph{Mathematics of Computation}, 83(288): 2033--2060.

\bibitem[{OpenAI(2023)}]{openai2023gpt4}
OpenAI. 2023.
\newblock GPT-4 Technical Report.
\newblock arXiv:2303.08774.

\bibitem[{Paranjape et~al.(2023)Paranjape, Lundberg, Singh, Hajishirzi,
  Zettlemoyer, and Ribeiro}]{paranjape2023art}
Paranjape, B.; Lundberg, S.; Singh, S.; Hajishirzi, H.; Zettlemoyer, L.; and
  Ribeiro, M.~T. 2023.
\newblock ART: Automatic multi-step reasoning and tool-use for large language
  models.
\newblock arXiv:2303.09014.

\bibitem[{Sap et~al.(2022)Sap, LeBras, Fried, and Choi}]{sap2022neural}
Sap, M.; LeBras, R.; Fried, D.; and Choi, Y. 2022.
\newblock Neural theory-of-mind? on the limits of social intelligence in large
  lms.
\newblock \emph{arXiv preprint arXiv:2210.13312}.

\bibitem[{Saxton et~al.(2019)Saxton, Grefenstette, Hill, and
  Kohli}]{saxton2019analysing}
Saxton, D.; Grefenstette, E.; Hill, F.; and Kohli, P. 2019.
\newblock Analysing Mathematical Reasoning Abilities of Neural Models.
\newblock arXiv:1904.01557.

\bibitem[{Shridhar, Stolfo, and Sachan(2023)}]{shridhar2023distilling}
Shridhar, K.; Stolfo, A.; and Sachan, M. 2023.
\newblock Distilling Reasoning Capabilities into Smaller Language Models.
\newblock arXiv:2212.00193.

\bibitem[{Strobelt et~al.(2022)Strobelt, Webson, Sanh, Hoover, Beyer, Pfister,
  and Rush}]{strobelt2022interactive}
Strobelt, H.; Webson, A.; Sanh, V.; Hoover, B.; Beyer, J.; Pfister, H.; and
  Rush, A.~M. 2022.
\newblock Interactive and visual prompt engineering for ad-hoc task adaptation
  with large language models.
\newblock \emph{IEEE transactions on visualization and computer graphics},
  29(1): 1146--1156.

\bibitem[{Wang et~al.(2022)Wang, Wei, Schuurmans, Le, Chi, Narang, Chowdhery,
  and Zhou}]{wang2022self}
Wang, X.; Wei, J.; Schuurmans, D.; Le, Q.; Chi, E.; Narang, S.; Chowdhery, A.;
  and Zhou, D. 2022.
\newblock Self-consistency improves chain of thought reasoning in language
  models.
\newblock \emph{arXiv preprint arXiv:2203.11171}.

\bibitem[{Wang(2002)}]{wang2002goldbach}
Wang, Y. 2002.
\newblock \emph{The Goldbach Conjecture}, volume~4.
\newblock World scientific.

\bibitem[{Wang(2022)}]{wang2022proof}
Wang, Y. 2022.
\newblock A Proof of Goldbach Conjecture by Mirror-Prime Decomposition.
\newblock \emph{WSEAS Transactions on Mathematics}, 21: 563--571.

\bibitem[{Wang and Chiew(2010)}]{wang2010cognitive}
Wang, Y.; and Chiew, V. 2010.
\newblock On the cognitive process of human problem solving.
\newblock \emph{Cognitive systems research}, 11(1): 81--92.

\bibitem[{Wei et~al.(2022)Wei, Wang, Schuurmans, Bosma, Xia, Chi, Le, Zhou
  et~al.}]{wei2022chain}
Wei, J.; Wang, X.; Schuurmans, D.; Bosma, M.; Xia, F.; Chi, E.; Le, Q.~V.;
  Zhou, D.; et~al. 2022.
\newblock Chain-of-thought prompting elicits reasoning in large language
  models.
\newblock \emph{Advances in Neural Information Processing Systems}, 35:
  24824--24837.

\bibitem[{Wu et~al.(2022)Wu, Jiang, Donsbach, Gray, Molina, Terry, and
  Cai}]{wu2022promptchainer}
Wu, T.; Jiang, E.; Donsbach, A.; Gray, J.; Molina, A.; Terry, M.; and Cai,
  C.~J. 2022.
\newblock Promptchainer: Chaining large language model prompts through visual
  programming.
\newblock In \emph{CHI Conference on Human Factors in Computing Systems
  Extended Abstracts}, 1--10.

\bibitem[{Yao et~al.(2023)Yao, Yu, Zhao, Shafran, Griffiths, Cao, and
  Narasimhan}]{yao2023tree}
Yao, S.; Yu, D.; Zhao, J.; Shafran, I.; Griffiths, T.~L.; Cao, Y.; and
  Narasimhan, K. 2023.
\newblock Tree of Thoughts: Deliberate Problem Solving with Large Language
  Models.
\newblock arXiv:2305.10601.

\bibitem[{Zhang et~al.(2022)Zhang, Zhang, Li, and Smola}]{zhang2022automatic}
Zhang, Z.; Zhang, A.; Li, M.; and Smola, A. 2022.
\newblock Automatic chain of thought prompting in large language models.
\newblock \emph{arXiv preprint arXiv:2210.03493}.

\bibitem[{Zhou et~al.(2022)Zhou, Muresanu, Han, Paster, Pitis, Chan, and
  Ba}]{zhou2022large}
Zhou, Y.; Muresanu, A.~I.; Han, Z.; Paster, K.; Pitis, S.; Chan, H.; and Ba, J.
  2022.
\newblock Large language models are human-level prompt engineers.
\newblock \emph{arXiv preprint arXiv:2211.01910}.

\end{thebibliography}

\end{document}